\documentclass[12pt]{elsarticle}
\usepackage[margin=1in]{geometry}

\usepackage{lscape}
\usepackage{verbatim}
\usepackage[symbol]{footmisc}

\usepackage{amssymb}
\usepackage{csquotes}
\usepackage{algorithm}
\usepackage{flushend}
\usepackage{balance}
\usepackage{enumitem}
\usepackage{adjustbox}
\usepackage{setspace}

\usepackage{booktabs}
\usepackage{fancyvrb}
\usepackage{rotating}
\usepackage{multirow}
\usepackage{dcolumn}
\usepackage{xcolor, soul}
\colorlet{yellowhl}{yellow!30}
\sethlcolor{yellowhl}
\usepackage[framemethod=tikz]{mdframed}

\usepackage{tikz,pgfplots}

\newcommand*\circled[1]{\tikz[baseline=(char.base)]{
            \node[shape=circle,draw,inner sep=1pt] (char) {#1};}}

\usepackage{caption}
\usepackage[noend]{algorithmic}

\usepackage{pifont}
\usepackage{qtree}
\usepackage{graphicx}
\usepackage{newfloat}
\usepackage[T1]{fontenc}

\usepackage{arydshln}

\DeclareFloatingEnvironment[fileext=lod]{diagram}

\newcommand{\specialcell}[2][c]{\begin{tabular}[#1]{@{}l@{}}#2\end{tabular}}

\usepackage{amsthm}

\usepackage{mathtools}

\usepackage{multirow}
\usepackage{dcolumn}
\usepackage{array}
\usepackage{lineno}

\usepackage{url}

\begin{document}

\begin{frontmatter}



\title{\large Improved Biomedical Word Embeddings in the Transformer Era}

\author[cs]{Jiho Noh}
\ead{jiho.noh@uky.edu}

\author[bmi,cs]{Ramakanth Kavuluru}
\ead{rvkavu2@uky.edu}

\address[bmi]{Division of Biomedical Informatics, Department of Internal Medicine, University of Kentucky}
\address[cs]{Department of Computer Science, University of Kentucky}



\begin{abstract}

\noindent Background: Recent natural language processing (NLP) research is dominated by neural network methods that employ word embeddings as basic building blocks. 
Pre-training with neural methods that capture local and global distributional properties (e.g., skip-gram, GLoVE) using free text corpora is often used to embed both words and concepts. Pre-trained embeddings are typically leveraged in downstream tasks using various neural architectures that are designed to optimize task-specific objectives that might further tune such embeddings.\\
\\
Objective: Despite advances in contextualized language model based embeddings, static word embeddings still form an essential starting point in BioNLP research and applications. They are useful in low resource settings and in lexical semantics studies. 
Our main goal is to build improved biomedical word embeddings and make them publicly available for downstream applications. \\
\\
Methods: We jointly learn word and concept embeddings by first using the skip-gram method and further fine-tuning them with correlational information manifesting in co-occurring Medical Subject Heading (MeSH) concepts in biomedical citations. This fine-tuning is accomplished with the transformer-based BERT architecture in the two-sentence input mode with a classification objective that captures MeSH pair co-occurrence.  We conduct evaluations of these tuned static embeddings using multiple datasets for word relatedness developed by previous efforts. \\
\\
Results: Both in qualitative and quantitative evaluations we demonstrate that our methods produce improved biomedical embeddings in comparison with other static embedding efforts. Without selectively culling concepts and terms (as was pursued by previous efforts), we believe we offer the most exhaustive evaluation of biomedical embeddings to date with clear performance improvements across the board. \\
\\
Conclusion: 
We repurposed a transformer architecture (typically used to generate dynamic embeddings) to improve static biomedical word embeddings using concept correlations.
We provide our code and embeddings for public use for downstream applications and research endeavors: \url{https://github.com/bionlproc/BERT-CRel-Embeddings}

\end{abstract}


\end{frontmatter}


\section{Introduction}

Biomedical natural language processing (BioNLP) continues to be a thriving field of research, garnering both academic interest and industry uptake. Its applications manifest across the full translational science spectrum. From extracting newly reported protein-protein interactions from literature to mining adverse drug events discussed in the clinical text, researchers have leveraged NLP methods to expedite tasks that would otherwise quickly become intractable to handle with a completely manual process. Computer-assisted coding tools such as 3M 360 Encompass, clinical decision making assistants such as IBM Micromedex with Watson, and information extraction API such as Amazon Comprehend Medical are popular use-cases in the industry.
As textual data explodes in the form of scientific literature, clinical notes, and consumer discourse on social media, NLP methods have become indispensable in aiding human experts in making sense of the increasingly data heavy landscape of biomedicine. The rise of deep neural networks (DNNs) in computer vision and NLP fields has quickly spread to corresponding applications in biomedicine and healthcare. Especially, as of now, BioNLP almost exclusively relies on DNNs to obtain state-of-the-art results in named entity recognition (NER), relation extraction (RE), and entity/concept linking or normalization (EN) --- the typical components in biomedical information extraction\footnote{Some exceptions exist when handling smaller datasets in highly specific domains where ensembles of linear models may prove to be better}.

\subsection{Neural word embeddings}
The central idea in DNNs for NLP is the notion of dense embeddings of linguistic units in $\mathbb{R}^d$ ($d$-dimensional vector space of real numbers) for $d$ that generally ranges from a few dozen to several hundreds. The unit is typically a word~\cite{bengio2003neural,collobert2008unified,mikolov2013efficient}, but can also be a subword~\cite{bojanowski2017enriching} (e.g., prefix/suffix) or even a subcharacter~\cite{yu2017joint} (for Chinese characters that can be broken down further). These dense embeddings are typically \textit{pre-trained} using large free text corpora (e.g., Wikipedia, PubMed citations, public tweets) by optimizing an objective that predicts local context or exploits global context in capturing distributional properties of linguistic units. Based on the well-known distributional hypothesis that words appearing in similar contexts are semantically related or share meaning~\cite{harris1954distributional}, this pre-training often leads to embeddings that exhibit interesting properties in $\mathbb{R}^d$ that correspond to shared meaning.  Once pre-trained, word embeddings are generally fine-tuned in a supervised classification task (with labeled data)  using a task-specific DNN architecture that builds on top of these embeddings.
While the notion of dense word embeddings existed  in the nineties (e.g., latent semantic indexing), neural embeddings together with task-specific DNNs have revolutionized the field of NLP over the past decade.

  A \textit{static word embedding} is a function that maps each unique word in a corpus to a single dense vector, which is fixed regardless of its use in the context. Word2vec~\cite{mikolov2013efficient} is one such method  that had an extensive influence on NLP applications due to its simple model architecture and efficient training techniques. Word2vec is a shallow neural network that predicts which words appear in the context of a target word (or vice versa). GloVe~\cite{pennington2014glove} extends the word2vec method and aims to approximate the word co-occurrence counts, with faster training and comparable performance even with a small corpus. GloVe differs in that word2vec is a learning-based predictive model, whereas GloVe is a count statistics-based model. FastText~\cite{bojanowski2017enriching} was an attempt to address OOV (Out-of-Vocabulary) problem by considering the representations of a word's constituent character-level $n$-grams. These models are the most representative static word embedding methods in NLP.

Since 2018, however, the static embeddings discussed thus far have been improved upon to address issues with polysemy and homonymy. Around the same time, transformers (such as BERT~\cite{devlin2019bert} and RoBERTa~\cite{liu2019roberta}), ELMo~\cite{peters2018deep}, and UMLFiT~\cite{howard2018universal}  have been developed to facilitate contextualized embeddings that generate the embedding of a word based on its surrounding context. This process typically generates different embeddings for polysemous occurrences of a word, such as when the word ``discharge'' is used to indicate bodily secretions or the act of releasing a patient from a hospital. Even for words that typically have a unique meaning, contextual embeddings might generate embeddings that more precisely capture the subtleties in how it is used in a particular context. Such contextualized embeddings might be better suited when predicting NER tags or composing word sequences toward a classification end-goal.

\subsection{Motivation for improved static embeddings}

Although contextualized embeddings are an excellent addition to the neural NLP repertoire, we believe there is merit in improving the static embeddings for various reasons: (1).~Contextualized models are based on language modeling and are more complex with multiple layers of recurrent units or self-attention modules. Base models tend to have tens of millions of parameters~\cite{rogers2020primer} and using them without GPUs in low-resource settings such as smart devices used in edge computing or IoT is infeasible. Simpler models that use static embeddings can be built with 1--2 orders of magnitude fewer parameters and can run on smaller CPUs even in low resource settings. While leaner transformers are  actively being investigated (e.g., DistilBERT~\cite{sanh2019distilbert}), they offer nowhere near the model size reduction needed for usage in low resource settings.  
Increasing use-cases of ``edge NLP''~\cite{edgenlp} further motivate our current effort. 
(2).~Static embeddings can be of inherent utility for linguists to continue to study lexical semantics of biomedical language by looking into word or subword embeddings and how they may be indicative of lexical relations (e.g., hypernymy and meronymy).  Another related use case is to study noun compound decomposition~\cite{kavuluru2012knowledge} in the biomedical language, which is typically treated as a bracketing task that ought to rely only on the local context within the noun compound. For example, \textit{candidate ((tumor suppressor) gene)}  and \textit{((tumor suppressor) gene) list} demonstrate two different decompositions of four-word compounds. (3).~Contextualized embeddings typically only make sense in languages that have large digitized corpora. For less known languages that have smaller repositories, the language modeling objective such embeddings rely on can lead to significant overfitting compared to static approaches~\cite{eisenschlos2019multifit}. (4).~Improved static word embeddings can also help initialize the embeddings before the process of language-modeling-based training ensues in the more expensive contextualized models\footnote{This clearly assumes that the same tokenization is appropriately maintained in both static and the subsequent contextualized models} to further enhance them (when compute power is not a major limitation).

\subsection{Related work~\label{sec:related-work}}
In this section, we briefly discuss previously proposed methods for training domain-specific word/concept embeddings, which we compare with our methods for this paper (as shown in Table~\ref{tbl:res-final}).
Wang et al.~\cite{wang2018comparison} trained word embeddings on unstructured electronic health record (EHR) data using fastText. The subword embeddings of the fastText model enabled them to obtain vector representations of OOV tokens.
Park et al.~\cite{park2019concept} proposed a model for learning UMLS concept embeddings from their definitions combined with corresponding Wikipedia articles~\cite{park2019concept}. The degree of relatedness between two concepts is measured by the cosine similarity between the corresponding concept vectors.
Zhang et al.~\cite{zhang2019biowordvec} proposed a similar method to ours for preparing the training corpus. They also used the MeSH RDF-based graph from which they sampled random paths to generate sequences of MeSH terms and used them to train word embeddings; in our work, we traverse the MeSH hierarchy to obtain single in-order path of MeSH concepts of which each node is represented by its preferred concept name, unique MeSH code, and its definition.
Yu et al.~\cite{yu2017retrofitting} also trained UMLS concept embeddings and fine-tuned them using a ``retrofitting'' method developed by Faruqui et al.~\cite{faruqui2015retrofitting}. They improved pre-trained embeddings using concept relationship knowledge defined in the UMLS semantic lexicon. Among different relationships, they claim that RO (\textit{has other relationship}) and RQ (\textit{related and possibly synonymous}) relationships returned the most improvements on the UMNSRS evaluation dataset.
Henry et al.~\cite{henry2019association} computed several association measures, such as \textit{mutual information}, with concept co-occurrence counts and measured the semantic similarity and relatedness between concepts. Overall, the Pearson's Chi squared association measure ($\chi^2$) performed the best.

\subsection{Overall contributions}
 {In this paper, we propose and evaluate methods to improve  static biomedical word embeddings to be made publicly available for downstream use by the community. Our main contributions follow.}
 \begin{itemize}
\item  {We jointly learn word and concept embeddings by leveraging definitional information for rare concepts to supplement concept-annotated corpora. 
Through this, we transfer bidirectional semantic signal between words and concepts, first using the PubTator concept-annotated corpus with fastText and subsequently using concept co-occurrences (along with their preferred names) to further fine-tune embeddings by adapting the BERT encoder in the two-sentence input mode.}
\item  {We assess the quality of the resulting embeddings with qualitative analyses and quantitative comparisons with those generated by prior methods on public datasets. Without selectively culling concepts and terms (as was pursued by previous efforts), we believe we offer the most exhaustive evaluation of static embeddings to date with clear performance improvements across the board. We provide our code and embeddings for public use for downstream applications and research endeavors:} \url{https://github.com/bionlproc/BERT-CRel-Embeddings}

\end{itemize}

\section{Methods}
Before we outline the  framework and intuition behind our  methods, we first motivate the idea of jointly learning embeddings for biomedical concepts and words in the context of our goals. 
Our framework toward improved biomedical word embeddings is depicted in Figure~\ref{fig:BMET-schema} whose components will be discussed in the rest of this section.

\subsection{High level intuition and overview}
\label{sec-high}

Biomedical concepts are analogous to named entities in general English. Names of genes, drugs, diseases, and procedures are typical examples of concepts. Just like entity linking in general NLP research, concept mapping is typically needed in BioNLP where concepts are to be mapped to their standardized counterparts in some expert curated terminology. This mapping part is harder in BioNLP given the variety of ways a concept can be referred to in running text. Often, there might not be much lexical overlap between different aliases that point to the same concept. For example, the procedure  \textit{ulnar collateral ligament reconstruction} is also called \textit{Tommy John surgery} and they both refer to the same medical subject heading (MeSH) concept code D000070638.
These aliases are provided in the corresponding terminology and the unified medical language system (UMLS) metathesaurus that integrates many such terminologies.

\begin{figure*}[htb]
    \centering
    \includegraphics[scale=0.5]{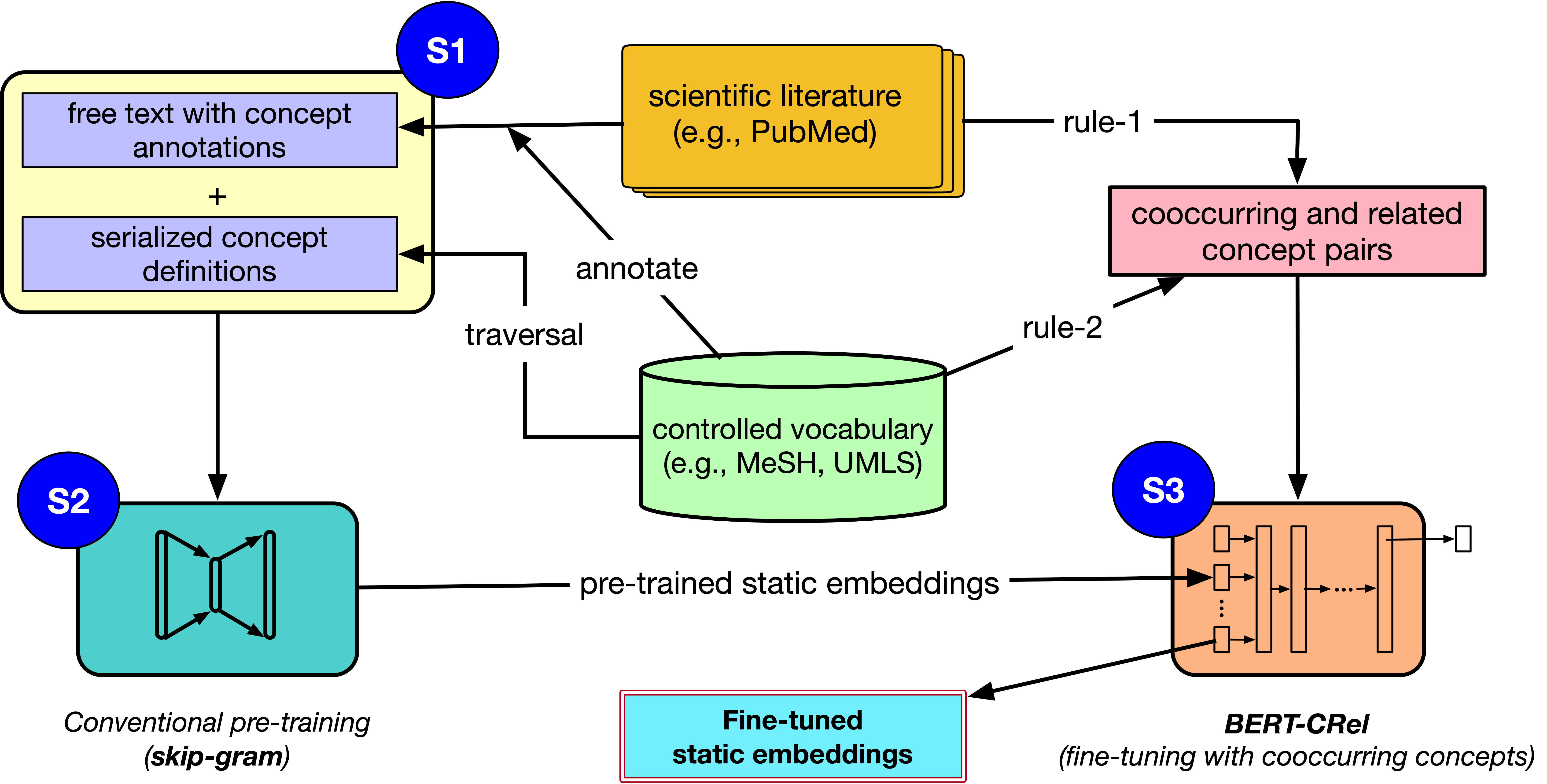}
    \caption{The schematic of our approach to improve word embeddings. S1 deals with pre-processing steps to create a concept enhanced corpus. S2 involves conventional pre-training using local context prediction objectives. S3 constitutes fine-tuning with distributional regularities based on co-occurrence. For S3, entity pairs are constructed based on two relevance rules: rule-1 is concept co-occurrence in a PubMed citation and rule-2 is proximity in a concept hierarchy~\label{fig:BMET-schema}}
\end{figure*}

Our first main idea is to use a well-known concept mapping tool to spot concepts in large biomedical corpora and insert those concept codes adjacent to the concept spans. This step is indicated as the \circled{S1} portion in Figure~\ref{fig:BMET-schema}.
Subsequently, run a pre-training method to embed both words and concepts in the same space in $\mathbb{R}^d$. This jointly learns embeddings for both words and concepts and enables two-way sharing of semantic signal: first  word embeddings are nudged to predict surrounding concepts, and as the pre-training window moves along the running text, concept embeddings are also nudged to predict neighboring words. In fact, this phenomenon has been exploited by multiple prior efforts~\cite{cai2018medical,choi2016multi,de2014medical} including in our prior work~\cite{sabbir2017knowledge}. Most of these efforts aim to learn concept embeddings that can be used in downstream applications. Here we demonstrate that this process also improves the word embeddings themselves. This process is indicated through the \circled{S2} part of Figure~\ref{fig:BMET-schema}. Our choice for biomedical concepts to be jointly learned is the set of nearly 30,000 MeSH codes that are used on a daily basis at the National Library of Medicine (NLM) by trained coders who assign 10--15 such codes per biomedical article.

On top of this joint pre-training approach, we introduce a novel application of the BERT encoder architecture to further fine-tune the word and concept embeddings with a classification objective that discriminates ``co-occurring'' MeSH codes (from PubMed citations) from random pairs of MeSH terms\footnote{ {We chose BERT style encoding instead of EMLo and ULMFiT because BERT's transformer architecture allows for a deeper sense of bidirectionality with the masked language modeling objective with multiple layers of attention; whereas ELMo and UMLFiT both use LSTMs, typically not amenable for parallelization with objectives that do not encode bidirectionality as well as BERT}}.  Here, co-occurrence refers to the two terms appearing in the same citation as determined by human coders who annotated it.
That is, the positive examples are derived from a set of MeSH codes assigned to a sampled biomedical citation, and negative examples are random pairs of MeSH codes from the full terminology. Intuitively, if two codes are assigned to the same article, they are clearly related in some thematic manner. Besides this, we also derive additional positive pairs from the MeSH hierarchy by choosing those that are separated by at most two hops. ``Jointness''  is incorporated here by appending each code with its preferred name. Specifically, in the two-sentence input mode for BERT, each sentence is a code and its preferred name appended next to it.  This code pair ``relatedness'' classification task further transfers signal between words and codes leading to demonstrable gains in intrinsic evaluations of resulting word embeddings.
These steps are captured through \circled{S3} in Figure~\ref{fig:BMET-schema}. We present more specifics and implementational details in Sections~\ref{sec:data} and \ref{sec:model}.

The resulting embeddings are evaluated for their semantic   representativeness using intrinsic evaluations with well-known datasets and also through qualitative analyses. The results show a substantial improvement in  evaluations compared to prior best approaches. Overall, we present an effective novel application of transformer architectures originally developed for contextualized embeddings to improve static word embeddings through joint learning and fine-tuning word/concept embeddings.

\subsection{Data Sources\label{sec:data}}
For S1 and S2 (in Figure~\ref{fig:BMET-schema}), to carry out conventional pre-training and learn word/concept embeddings, we seek a free publicly available resource that comes with annotations of biomedical concepts from a well-known terminology. This is readily made available through the PubTator~\cite{wei2019pubtator} initiative from BioNLP researchers at the NLM. It has over 30 million PubMed citations (abstracts and titles from the 2021 baseline) and over 3 million full-text articles with high-quality annotations for genes (and their variants), diseases, chemicals, species, and cell lines. Our choice for the concept vocabulary was MeSH (2021 version) because the diseases and chemicals from PubTator have mappings to MeSH codes; furthermore, with nearly 30K concepts, MeSH is fairly representative of the general concept space in biomedicine. Additionally, MeSH concepts also come with brief definitional blurbs describing their meaning in general-purpose English (more later). We use these blurbs in pre-training for MeSH concepts that do not appear in PubTator annotations.

\subsubsection{Concept annotated corpus for pre-training \label{sec-pret}}

Pre-training step \circled{S2} in Figure~\ref{fig:BMET-schema} uses fastText~\cite{bojanowski2017enriching} for training static embeddings.
FastText improves upon the basic skip-gram model by learning word embeddings as compositions of constituent character n-grams and their representations. The corpus for this is a sample subset (1--2\%) of the PubTator dataset such that each PubMed citation sampled contains at least two annotations with MeSH concepts. MeSH codes from the annotations are inserted immediately after the corresponding concept spans in texts. To distinguish MeSH codes from regular words, we represent them as \texttt{ConceptCode}||\texttt{SourceVocab}, essentially a concatenation of the concept code and \texttt{SourceVocab}, an abbreviation for the source terminology. Although MeSH codes are unique enough, we chose this formatting to be amenable to a general setup with multiple terminologies. With this, consider the example title: ``A multi-centre international study of salivary hormone oestradiol and progesterone measurements in ART monitoring.'' With the corresponding codes inserted, this title is transformed into: \textit{A multi-centre international study of salivary hormone oestradiol \texttt{D004958MeSH} and progesterone \texttt{D011374MeSH} measurements in ART monitoring}. The two codes inserted next to ``oestradiol'' and ``progesterone'' were identified by PubTator.

Our goal is to imbue a two-way semantic signal between all types of concepts and related words. However, only a portion of the MeSH headings (9,477 out of 29,915) is referred to in the PubTator annotations. Hence, we ought to supplement PubTator based training data with additional texts that contain the missing MeSH codes. This is where we exploit the definitional information of concepts provided by MeSH creators. With this, each MeSH concept provides a textual snippet for fastText. The snippet supplied is the concatenation of  the preferred name, source code, and definition of the concept. For example, the MeSH code D008654 for the concept Mesothelioma results in the textual input: ``\textbf{Mesothelioma} \texttt{D008654MeSH}  \textit{A tumor derived from mesothelial tissue (peritoneum, pleura, pericardium). It appears as broad sheets of cells, with some regions containing spindle-shaped, sarcoma-like cells and other regions showing adenomatous patterns. Pleural mesotheliomas have been linked to exposure to asbestos.}''
This means, for codes that may never show up in any annotated PubTator documents, we guarantee a single document  that is constructed in this manner tying the concept with words that are highly relevant to its meaning. These are the ``serialized concept definitions'' referred to in the \circled{S1} component of Figure~\ref{fig:BMET-schema}.  These additional documents are supplied in an in-order traversal sequence of the MeSH hierarchy to fastText as a ``mega'' document where adjacent documents correspond to hierarchically related concepts. Table~\ref{tbl:fastText_data_stats} describes the statistics of the textual resources used for pre-training step \circled{S2}.

\begin{table}[hbt]
  \centering
  \caption{ {Descriptive statistics of training examples for pre-training initial static embeddings}}
  \label{tbl:fastText_data_stats}
  \begin{tabular}{@{}rl@{}}
    \toprule
    \textit{PubTator} & 
      \begin{tabular}[t]{@{}l@{}}
        Total \# of sampled documents: 501,639 / 30,017,978 ($\sim$2\%) \\
        Total \# of unique MeSH descriptors covered: 9,477 \\
        Average \# of entity mentions per document: 7.32 \\
        Average \# of tokens per document: 204.12
      \end{tabular} \\
    \midrule
    \textit{MeSH} & 
      \begin{tabular}[t]{@{}l@{}}
        Total \# of entities (descriptors only): 29,915 \\
        Average \# of tokens per entity name: 2.00 \\
        Average \# of tokens per entity definitional information 28.80
      \end{tabular} \\
    \bottomrule
  \end{tabular}
\end{table}

\subsubsection{Training examples for code pair relatedness classification~\label{sec:data-BMET-bert}}
Component \circled{S3} of Figure~\ref{fig:BMET-schema} involves model BERT-CRel to further fine-tune  word and concept embeddings by capturing concept relatedness (CRel). It is a canonical transformer~\cite{vaswani2017attention} model for a binary classification task. In essence, this is repurposing the BERT architecture without any pre-training for the language modeling objective; we retain  the classification objective with an additional feedforward layer and sigmoid unit feeding off of the \texttt{[CLS]} token output. The input is a pair
($m^i$, $m^j$) of ``related'' MeSH concepts in the two-sentence input mode following the format

\begin{center}
  \texttt{[CLS]} $m^i \, w^i_1 \cdots w^i_n$ \texttt{[SEP]} $m^j \, w^j_1 \cdots w^j_m$ \texttt{[SEP]}
\end{center}
where $m^{i}$ and $m^j$ are related MeSH codes  and $w^{i}_1 \cdots w^{i}_{n}$ is the \textit{preferred name} of $m^i$.   \texttt{[CLS]} and \texttt{[SEP]} are well-known special tokens used in BERT models.

Positive training pairs ($m^i$, $m^j$) are generated using two rules. Rule-1 deems the pair to be related if both codes were assigned to some document in the sample corpus $C$ by coders at the NLM. More formally, the set of all such positive pairs
\[ R_{C} = \bigcup_{c \in C}  \, \{ (m^i, m^j):  \forall_{i \neq j} \, m^i, m^j \in \mathcal{M}(c)  \}, \]
where $\mathcal{M}(c)$ is the set of MeSH concepts assigned to citation $c$.
Rule-2 considers a pair to be related if the codes are connected by at most two hops in the directed-acyclic MeSH graph $G_{MeSH}$.
These would capture parent/child, grand parent/child, and sibling connections between concepts.
Specifically,
\[ R_{MeSH} = \{ (m^i, m^j): d^{G_{MeSH}}(m^i, m^j) \leq 2, \, \forall_{i \neq j} \,m^i, m^j \in G_{MeSH} \}  \cup R^{MeSH}_{SA} \cup R^{MeSH}_{PA}, \]

where $d$ is graph distance,  $R^{MeSH}_{SA}$ is the set of ``see also'' relations, and $R^{MeSH}_{PA}$ is the set of  ``pharmacological action'' relations defined between MeSH concepts by the NLM. These auxiliary relations are not part of the MeSH hierarchy but are publicly available to mine. For instance, the concept \textit{Multiple Myeloma} has a see-also link to the concept \textit{Myeloma Proteins}, which in turn has a pharm-action connection to the concept \textit{Immunologic Factors}.
It is not difficult to see that these relations also capture strong semantic relatedness between concepts.
$R_C \cup R_{MeSH}$ is the full set of positive relations used to fine-tune word/concept embeddings with BERT-CRel.
To generate the same number of negative examples, we randomly sample the MeSH concept pairs across the entire vocabulary, retaining the term frequency distribution.
Details of the numbers of examples used are in Table~\ref{tbl:crel_data_stats}.

\begin{table}[hbt]
  \centering
  \caption{ {Descriptive statistics of training examples for the BERT-CRel component}}
  \label{tbl:crel_data_stats}
  \begin{tabular}{@{}rl@{}}
    \toprule
    \textit{PubTator} &
      \begin{tabular}[t]{@{}l@{}}
      Total \# of sampled documents: 136,437 ($\sim$0.5\%) \\
      Total \# of unique MeSH descriptors covered: 24,995 \\
      Total \# of concept pairs generated (including negative pairs): 8,752,116 \\
      \end{tabular}\\
    \midrule
    \textit{MeSH} &
      \begin{tabular}[t]{@{}l@{}}
      Total \# of concept pairs generated: 363,697 \\
      Total \# of concept pairs generated from auxiliary relationships: 23,598\\
      \end{tabular}\\
    \bottomrule
  \end{tabular}
\end{table}

\subsection{Models and Configurations~\label{sec:model}}

\subsubsection{fastText$^{+}$: adjustments to fastText for word/concept pre-training \label{sec-fastmod}}

As indicated in Section~\ref{sec-pret} we use fastText~\cite{bojanowski2017enriching} for the initial pre-training on the concept-annotated corpus created through PubTator and MeSH definitional information. Building on the skip-gram model~\cite{mikolov2013efficient}, fastText additionally models and composes character n-grams to form word embeddings, thus accounting for subword information. This can capture relatedness among morphological variants and in exploiting regularities in lexical meaning manifesting in word forms through suffixes, prefixes, and other lemmata. It also helps in forming better embeddings on the fly for some unseen words (through the constituent character n-grams) instead of relying on the catch-all UNK embeddings that are typically used.
However, we do not want this subword decomposition to occur when dealing with concept embeddings because they are atomic units, and there is no scope for unseen tokens given we know the full code set upfront. Hence we impose the following two constraints.
\begin{enumerate}
  \item Concept codes (e.g., D002289MeSH) are not   decomposed into subword vectors; the model thus is forced to recognize the concept codes from the corresponding tokens by the unique format \texttt{ConceptCode}||\texttt{SourceVocab}.
  \item The output vocabulary must contain the full set of concept codes (here, MeSH descriptors) regardless of their frequencies in the corpus unlike the default case where fastText imposes a minimum frequency for character n-grams.
\end{enumerate}

For the full implementation details of fastText, we refer to the original paper by Bojanowski et al.~\cite{bojanowski2017enriching}. Here, we only highlighted the modifications we sought to handle concept tokens. This adapted version of fastText is henceforth called fastText$^{+}$ in this paper.
Table~\ref{tbl:bmet-ft-params} lists the empirically chosen hyperparameters for training fastText for our concept-annotated corpus. Note that the dimensionality of word vectors (\textit{dim}) is intentionally chosen to be divisible by 12, the number of transformer blocks in the subsequent fine-tuning phase through the BERT architecture.
\begin{table}[htb]
\renewcommand{\arraystretch}{1.05}
    \caption{Hyperparameters for word/concept pre-training through fastText~\label{tbl:bmet-ft-params}}
    \centering
    \begin{tabular}{@{}lr@{}}
        \toprule
        Parameters & Values \\
        \midrule
        \emph{minCount} (required number of word occurrences) & 5 \\
        \emph{dim} (dimensionality of word vectors) & 396 \\
        \emph{ws} (size of context window) & 30 \\
        \emph{epoch} (number of epochs) & 5 \\
        \emph{minn} (min.~length of character ngrams) & 3 \\
        \emph{maxn} (max.~length of character ngrams) & 6 \\
        \bottomrule
    \end{tabular}
\end{table}

\subsubsection{BERT-CRel: Fine-tuning static embeddings with the concept relatedness objective}

\begin{figure}[htb]
    \centering
    \includegraphics[scale=0.60]{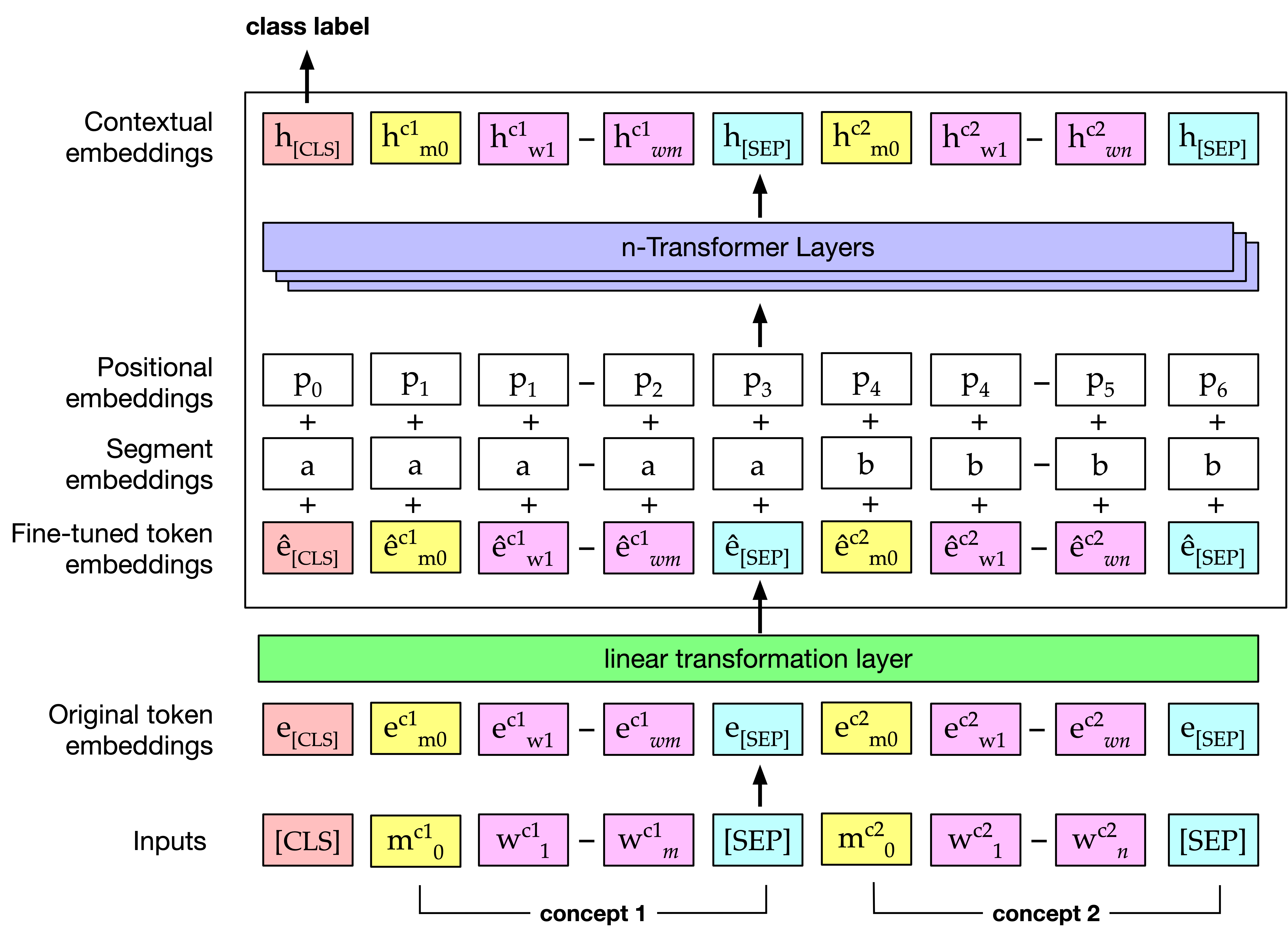}
    \caption{BERT-CRel concept relatedness classification model to fine-tune embeddings~\label{fig:BMET-bert-schema}}
\end{figure}

We introduced BERT-CRel in Section~\ref{sec:data-BMET-bert} to further fine-tune pre-trained word/concept embeddings learned with fastText$^{+}$.  BERT-CRel is a shallow transformer encoder, which reads the textual representations of a concept pair and predicts their relatedness as a binary classification task. Note that is unlike the original purpose of BERT --- to build contextualized embeddings. Furthermore, we do not use any pre-trained BERT model (such as SciBERT) because our framework does not suit the \textit{WordPiece} tokenization that is typically used. What is available at this stage are the pre-trained word/concept embeddings from fastText$^{+}$. So we repurpose BERT as shown in Figure~\ref{fig:BMET-bert-schema}. Here we apply a linear transformation on the initial pre-trained static embeddings.

The input texts are tokenized using a simple white space-based split function followed by a text clean-up process. Initially, we load the original token embeddings with the pre-trained static embeddings  from fastText$^{+}$. We provide examples of concept pairs (as outlined in Section~\ref{sec:data-BMET-bert}) along with their binary relatedness labels to the model. Each input sequence starts with \texttt{[CLS]}, followed by a pair of concept phrases (code token followed by the preferred name for each concept) separated by \text{[SEP]}. While training, the first \texttt{[CLS]} token collects all the features for determining the relatedness label between two concepts. We add a linear transformation layer following the original token embeddings to apply subtle adjustments to the given token embeddings. This linear layer is initialized with the identity matrix.

\subsubsection*{Two-step optimization}
We take a two-step optimization approach where during the first step, we focus on optimizing the classification model before fine-tuning the pre-trained embeddings. 
To accomplish this, during the first step, only the transformer layers are updated with the specified range of learning rates $[lr_{max}^\alpha, lr_{min}^\alpha]$, starting with $lr_{max}^\alpha$ and decreasing with time. Once the optimizer reaches the minimum learning rate ($lr_{min}^\alpha$), we initiate the next optimization schedule by applying another range of learning rates $[lr_{max}^\beta, lr_{min}^\beta]$ and start computing gradients of the linear transformation layer. This new range is to update the linear transformation layer~($\Theta$) and the pre-trained embeddings from fastText$^{+}$~($E$).

This second step is implemented using   multi-stage annealing within learning rate range  $[lr_{max}^\beta, lr_{min}^\beta]$. That is, we first update the linear layer  with fixed embeddings from the previous stage. This stops when the learning rate decreases to $lr_{min}^\beta$. At this point, the embeddings are updated ($E_{i+1} = \Theta_{i} E_{i}$) at once using the state of the parameters and $\Theta_{i+1}$ is set back to $I$ (identity matrix). The learning rate is then reset to a higher value that starts at  $ lr_{i+1} = \gamma^{i+1} \cdot lr_{max}^\beta$ ($\gamma < 1$); and the process of updating $\Theta_{i+1}$ continues with fixed $E_{i+1}$. This alternating process of freezing $E$ and updating $\Theta$ and then updating $E$ after reaching minimum learning rate  is repeated until $lr_{i+1}$ reaches  $lr_{min}^\beta$ (which is the default manner in which PyTorch's \textit{ReduceLRonPlateau} operates). $E_1$ is the pre-trained set of embeddings from fastText$^{+}$ and $\Theta_1$ is initialized with $I$. Intuitively, this lets the learning rate bob within the $[lr_{max}^\beta, lr_{min}^\beta]$ range inspired by cyclical learning rate schedules~\cite{smith2017cyclical}  designed to overcome saddle point plateaus.

  {Intuitively, this two-stage optimization approach is to ensure that the layer closer to the final layer is trained first in isolation (by freezing other layers) to give it ample scope to adapt to the final objective. This helps the model better leverage the word embeddings that were pre-trained via fastText. This freezing of the embedding layer allows it to exert sustained influence on the fine-tuning of the classification layers without undergoing any catastrophic forgetting. Once this happens, all layers are unfrozen in the end to be trained a final time for better end-to-end convergence. This process has been termed ``chain thaw'' and was shown to work better compared to conventional training methods}~\cite{felbo2017using}.

\subsubsection*{Implementation details~\label{sec:implementation-details}}

We use PyTorch and HuggingFace's \textit{BertForSequenceClassification} model to implement BERT-CRel. The model is evaluated on the validation set every 10,000 steps. Binary cross-entropy is the loss function used. We save the improved word embeddings of the best model according to the UMNS dataset (more later) evaluation results. We use \textit{ReduceLRonPlateau} with the initial learning rate $lr_{max}^\alpha=\texttt{3e-5}$ and the minimum learning rate $lr_{min}^\alpha= \texttt{2e-5}$ with decay $\gamma = 0.9$ for the initial step of updating just the transformer layers. The scheduler reduces learning rates by $\gamma$ once it sees no improvement on the validation results three consecutive times. While fine-tuning static embeddings, during the multi-stage annealing process, we set the learning rates from \texttt{3e-5} ($lr_{max}^\beta$) to \texttt{1e-5} ($lr_{min}^\beta$) with $\gamma = 0.8$.  The values for $\gamma$s are empirically chosen with the intention of providing slower learning rate reduction for the initial model learning than the fine-tuning process.

\subsection{Evaluation Scenarios~\label{sec:experiments}}

\subsubsection{Qualitative evaluations}

As a qualitative evaluation, we examine the representation learning quality of the embeddings produced by BERT-CRel. This is done in the context of other prior approaches for generating  biomedical word embeddings. For the sake of comparison, we use the same set of biomedical query terms (usually noun phrases) used in Wang et al.'s study~\cite{wang2018comparison}. The task is to retrieve five \textit{closest} terms in the word/concept embedding space to each query term and assess how related they actually are to the query term. For example, given the word `aspirin,' we expect to see related terms such as `blood thinner', `anti-inflammatory drug', or  `clopidogrel' (shares functionality with aspirin). These typically include hyponyms, hypernyms, or co-hyponyms.
Besides terms by Wang et al.~\cite{wang2018comparison}, we also examine the neighbors of most popular acronyms used in biomedical literature; we find up to five closest terms to the acronym and the corresponding MeSH codes. We used two available algorithms for acronym extraction, the Schwartz and Hearst algorithm~\cite{schwartz2002simple} and  ALICE~\cite{ao2005alice}, and obtained 331 most frequently used acronyms in the PubMed citations for this purpose.
We note that for multi-word terms, we simply take the average of constituent word embeddings before retrieving the closest words and concepts.

\subsubsection{Quantitative evaluations~\label{sec:quant-evals}}
Intrinsic evaluations for word embeddings examine the quality of representativeness that is independent of downstream tasks. We use publicly available reference datasets for measuring the relatedness between biomedical concepts. With the reference standards, we can evaluate the quality of vector representations for computing relatedness between biomedical terms compared to human judgments.
 Each instance within a dataset consists of a pair of biomedical concepts and the corresponding relatedness score judged by human experts such as physicians and medical coders. Some of the datasets also provide corresponding UMLS concept codes. The terms that occur in these datasets are more often seen in the biomedical domains than in other fields. Table~\ref{tbl:intr-eval-sets} enumerates the reference datasets we use, where the middle column indicates the number of concept pairs within each dataset.

\begin{table}[htb]
    \centering
    \renewcommand{\arraystretch}{1.05}
    \caption{Datasets of biomedical concept pairs for similarity/relatedness evaluations.~\label{tbl:intr-eval-sets}}
    \small
    \begin{tabular}{@{}lrlc@{}}
        \toprule
        Dataset name (alias) & Size  & Judged by \\
        \midrule
        UMNSRS-Sim (\texttt{UMNS})~\cite{pakhomov2010semantic} & 566 & medical residents \\
        UMNSRS-Rel (\texttt{UMNR})~\cite{pakhomov2010semantic} & 587 & medical residents \\
        MayoSRS (\texttt{MAYO})~\cite{pakhomov2018semantic} & 101 & physicians and coders \\
        MiniMayoSRS (\texttt{MMY[P/C]})~\cite{pedersen2007measures} & 29 & physicians and coders \\
        Pedersen's (\texttt{PDS[P/C]})~\cite{pedersen2007measures} & 30 & physicians \\
        Hliaoutakis' (\texttt{HLTK})~\cite{hliaoutakis2005semantic} & 36 & mostly physicians \\
        \bottomrule
    \end{tabular}
\end{table}

We expand the instances by linking the concepts to corresponding MeSH codes. We utilize the UTS (UMLS Terminology Services) API\footnote{\url{https://documentation.uts.nlm.nih.gov}} to find the most similar MeSH codes to the concepts. When available, we exploit the UMLS codes provided along with the datasets; otherwise, we query by the concept name.
We use the cosine vector similarity to measure the semantic match between two concepts/terms.
Here also, if the concept name is composed of multiple words, we take the mean vector of its constituent word representations. If the word is  OOV (Out-of-Vocabulary), the \texttt{[UNK]} token vector learned in BERT-CRel training process is used. If \texttt{[UNK]} token is not available, for the fastText$^{+}$ pre-trained embeddings, we assume the relatedness score of the pair to be $0$ as default.  Finally, a ranked list of concept pairs based on cosine scores is compared against the ground truth expert ranking using the Spearman's rank correlation coefficient $\rho$.

We test the significance of performance gains from our best method relative to the best prior score on different datasets. Given correlation coefficients ($\rho$'s) are not normally distributed and the actual number of  examples used are different, we use the \textit{Fisher Z-transformation} and the one-tailed p-values for comparison~\cite{fisher1915frequency}. We compute the normalized means,
\begin{equation*}
  \centering
  \mu_1' = \tanh^{-1}(\rho_1) = \frac{1}{2}\ln \left(\frac{1+\rho_1}{1-\rho_1}\right), \qquad
  \mu_2' = \tanh^{-1}(\rho_2) = \frac{1}{2}\ln \left(\frac{1+\rho_2}{1-\rho_2}\right)
\end{equation*}
We compute $z$-score,
\begin{equation*}
  z = \frac{\mu_1' - \mu_2'}{S} \sim N(0, 1),
\end{equation*}
with the standard error of the difference between means where $n$'s are the sample sizes:
\begin{equation*}
  S = \sqrt{S_1^2 + S_2^2} = \sqrt{\frac{1}{n_1-3} + \frac{1}{n_2-3}}.
\end{equation*}
The p-value is computed using this Z-score.

\begin{sidewaystable*}
  \caption{Five most similar terms to selected biomedical concepts trained from different models and textual resources, MeSH names: (i) Diabetes Melitus (ii) Diabetes Mellitus, Type 2 (iii) Ulcer (iv) Peptic Ulcer (v) Stomach Neoplasms (vi) Colorectal Neoplasms (vii) neoplasms (viii) Dyspnea (ix) Pharyngeal Diseases (x) Opioid-Related Disorders (xi) Aspirin}
  \label{tbl:five-similar-words}
    \small
    \begin{adjustbox}{width=\textwidth}
    \begin{tabular}{@{}lllllll@{}}
        \toprule
        Query term & \textbf{fastText$^{+}$} (PubMed) & \textbf{BERT-CRel} (PubMed) & Wang et al.'s (EHR) & Wang et al.'s (PMC) & GloVe (Wiki+Giga) & W2V (Google News) \\
        \midrule
        diabetes
        & D003920 \textsuperscript{i} & D003920 \textsuperscript{i} & mellitus & cardiovascular & hypertension & diabetics \\
        & mellitus & mellitus & uncontrolled & nonalcoholic & obesity & hypertension \\
        & nondiabetes & nondiabetes & cholesterolemia & obesity & arthritis & diabetic \\
        & diabetic & D003924 \textsuperscript{ii} & dyslipidemia & mellitus & cancer & diabetes\_mellitus \\
        & D003924 \textsuperscript{ii} & diabetic & melitis & polycystic & alzheimer & heart\_disease \\
        \midrule
        peptic ulcer disease
        & D014456 \textsuperscript{iii}  & D014456 \textsuperscript{iii} & scleroderma & gastritis & ulcers & ichen\_planus \\
        & ulcers  & D010437 \textsuperscript{iv} & duodenal & alcoholism & arthritis & Candida\_infection \\
        & D010437 \textsuperscript{iv} & ulcers & crohn & rheumatic & diseases & vaginal\_yeast\_infections \\
        & gastroduodenitis &  D013274 \textsuperscript{v} & gastroduodenal & ischaemic & diabetes & oral\_thrush \\
        & ulceration & gastroduodenitis & diverticular & nephropathy & stomach & dermopathy \\
        \midrule
        colon cancer
        & colorectal & D015179 \textsuperscript{vi} & breast & breast & breast & breast \\
        & D015179 \textsuperscript{vi} & colorectal & ovarian & mcf & prostate & prostate \\
        & cancers & cancers & prostate & cancers & cancers & tumor \\
        & D009369 \textsuperscript{vii}  & colorectum & postmenopausally & tumor\_suppressing & tumor & pre\_cancerous\_lesion \\
        & colorectum & D009369 \textsuperscript{vii} & caner & downregulation & liver & cancerous\_polyp \\
        \midrule
        dyspnea
        & D004417 \textsuperscript{viii} & D004417 \textsuperscript{viii} & palpitations & sweats & shortness & dyspnoea \\
        & dyspnoea & dyspnoea & orthopnea & orthopnea & breathlessness & pruritus \\
        & shortness & shortness & exertional & breathlessness & cyanosis & nasopharyngitis \\
        & breathlessness & breathlessness & doe & hypotension & photophobia & symptom\_severity \\
        & dyspnoeic & dyspnoeic & dyspnoea & rhonchi & faintness & rhinorrhea \\
        \midrule
        sore throat
        & pharyngitis & pharyngitis & scratchy & runny & shoulder & soreness \\
        & throats & D010608 \textsuperscript{ix}  & thoat & rhinorrhea & stomach & bruised \\
        & pharyngolaryngitis & pharyngolaryngitis & cough & myalgia & nose & inflammed \\
        & tonsillopharyngitis & pharyngotonsillitis & runny & swab\_fecal & chest & contusion \\
        & rhinopharyngitis & rhinopharyngitis & thraot & nose & neck & sore\_triceps \\
        \midrule
        low blood pressure
        & pressures & pressures & readings & dose & because & splattering\_tombstones \\
        & hemodynamics & flow & pressue & cardio\_ankle & result & Zapping\_nerves\_helps \\
        & subpressure & arterial & presssure & ncbav & high & pressue \\
        & arterial & high & bptru & preload & enough & Marblehead\_Swampscott\_VNA \\
        & normotension & hemodynamics & systolically & gr & higher & pill\_Norvasc \\
        \midrule
        opioid
        & opioids & opioids & opiate & opioids & analgesic & opioids \\
        & opiate & opiate & benzodiazepine & nmda\_receptor & opiate & opioid\_analgesics \\
        & nonopioid & nonopioid & opioids & affective\_motivational & opioids & opioid\_painkillers \\
        & nonopioids & morphine & sedative & naloxone\_precipitated & anti-inflammatory & antipsychotics \\
        & D009293 \textsuperscript{x} & nonopioids & polypharmacy & hyperlocomotion & analgesics & tricyclic\_antidepressants \\
        \midrule
        aspirin
        & D001241 \textsuperscript{xi} & D001241 \textsuperscript{xi} & ecotrin & chads & ibuprofen & dose\_aspirin \\
        & acetylsalicylic & acetylsalicylic & uncoated & vasc & tamoxifen & ibuprofen \\
        & nonaspirin & nonaspirin & nonenteric & newer & pills & statins \\
        & aspirinate & aspirinate & effient & cha & statins & statin \\
        & aspirinated & antiplatelet & onk & angina & medication & calcium\_supplements \\
        \bottomrule
    \end{tabular}
    \end{adjustbox}
\end{sidewaystable*}

\begin{sidewaystable*}[p]
    \small
    \caption{Nearest neighbors of most common biomedical abbreviations in the BMET-CRel trained embeddings}
    \label{tbl:res-acronyms}
    \begin{adjustbox}{width=\textwidth}
    \centering
    \begin{tabular}{@{}lll@{}}
        \toprule
        \specialcell{Acronyms} & Close to Word & Close to Code \\
        \midrule
        \specialcell{MRI\\(MeSH: D008279\\Name: Magnetic Resonance Imaging)} &
        \specialcell{imaging\\mris\\weighted\\tesla\\magnetic} &
        \specialcell{
          D066235 (Fluorine-19 Magnetic Resonance Imaging) \\
          D038524 (Diffusion Magnetic Resonance Imaging) \\
          D000074269 (Resonance Frequency Analysis) \\
          D000081364 (Multiparametric Magnetic Resonance Imaging) \\
          D017352 (Echo-Planar Imaging) }\\
        \midrule
        \specialcell{BMI\\(MeSH: D015992\\Name: Body Mass Index)} &
        \specialcell{overweight\\waist\\circumference\\whr\\D009765 (Obesity)} &
        \specialcell{
          D065927 (Waist-Height Ratio) \\
          D049629 (Waist-Hip Ratio) \\
          D049628 (Body Size) \\
          D064237 (Lipid Accumulation Product) \\
          D001823 (Body Composition) }\\
        \midrule
        \specialcell{CT\\(MeSH: D014057\\Name: Computed Tomography)} &
        \specialcell{scans\\tomographic\\computed\\scan\\tomography} &
        \specialcell{
          D014056 (Tomography, X-Ray) \\
          D055114 (X-Ray Microtomography) \\
          D000072078 (Positron Emission Tomography Computed Tomography) \\
          D055032 (Electron Microscope Tomography) \\
          D014055 (Tomography, Emission-Computed) }\\
        \midrule
        \specialcell{NO\\(MeSH: D009569\\Name: Nitric Oxide)} &
        \specialcell{significant\\any\\did\\not\\both} &
        \specialcell{nitric\\oxide\\inos\\nos\\D013481 (Superoxides) } \\
        \midrule
        \specialcell{ROS\\(MeSH: D017382\\Name: Reactive Oxygen Species)} &
        \specialcell{
          D017382 (Reactive Oxygen Species) \\
          oxidative \\
          h2o2 \\
          oxidant \\
          D013481 (Superoxides)} &
        \specialcell{
          ros \\
          oxidative \\
          h2o2 \\
          D006861 (Hydrogen Peroxide) \\
          D013481 (Superoxides) } \\
        \midrule
        \specialcell{PCR\\(MeSH: D016133\\Name: Polymerase Chain Reaction)} &
        \specialcell{polymerase\\qpcr\\primers\\taqman\\rt} &
        \specialcell{
          D054458 (Amplified Fragment Length Polymorphism Analysis) \\
          D020180 (Heteroduplex Analysis) \\
          D022521 (Ligase Chain Reaction) \\
          D060885 (Multiplex Polymerase Chain Reaction) \\
          D024363 (Transcription Initiation Site) } \\
        \midrule
        \specialcell{AD\\(MeSH: D000544\\Alzheimer Disease)} &
        \specialcell{ D000544 (Alzheimer Disease) \\
          alzheimer\\alzheimers\\abeta\\dementias } &
        \specialcell{alzheimer\\alzheimers\\ad\\abeta\\D003704 (Dementia)} \\
        \bottomrule
    \end{tabular}
    \end{adjustbox}
\end{sidewaystable*}

\section{Results and Discussion~\label{sec:results}}

\subsection{Qualitative evaluation}

We first discuss observations from the qualitative assessments conducted.
Table~\ref{tbl:five-similar-words} shows the five most related terms to a given biomedical term across several available embeddings. Sample query terms are in three groups: disease name, symptoms, and drug names. In the table, the fastText$^{+}$ column denotes the results obtained from the pre-trained static embeddings with the joint learning of word and concept embeddings (Section~\ref{sec-fastmod}). The BERT-CRel column indicates the results obtained from the improved static embeddings by the concept-relatedness classification task with the BERT encoder model.
We notice that both of our approaches (fastText$^{+}$ and BERT-CRel) surface a coherent set of words and concepts related to the query terms. Also, corresponding MeSH codes returned allow us to interpret input terms in an indirect but more precise way. For example, D015179 (Colorectal Neoplasms) exactly matches the query term ``colon cancer'' while other words are indicating relevant words but may not be as specific (e.g., ``cancers''). The returned words for the query term ``sore throat'' also demonstrate better ability in finding related terms. We were able to retrieve specific related disease names  such as \textit{pharyngitis}, \textit{pharyngolaryngitis}, and \textit{rhinopharyngitis}. The more primitive methods do not produce terms that are as tightly linked with the theme conveyed by query terms compared with our methods.
Between our fastText$^{+}$ and BERT-CRel rankings, there is a non-trivial overlap of terms, but the relative order seems to have changed due to the fine-tuning process. We see more examples  where BERT-CRel  ranks MeSH codes that precisely match the query term higher than the fastText$^{+}$ ranking. Also, BERT-CRel appears to surface  related terms that are not just morphological variants of the query term. For example, for the ``opioid'' query, it returns morphine, which is not returned in any other methods. However, other methods also seem to surface some interesting related terms such as ``analgesics'', a broader term that refers to pain relievers.

Table~\ref{tbl:res-acronyms} shows the mapping between  some commonly used biomedical acronyms and their nearest terms; the second column lists terms that are close to the acronym, and the third column contains terms close to the corresponding MeSH code. The results in the third column show how the distributional representations of MeSH codes are affected by the training sources. As mentioned earlier, PubTator annotates biomedical concepts that only belong to the following categories: gene, mutation, disease names, chemical substances, and species. Consequently, the MeSH codes for some acronyms (e.g., MRI, BMI, CT, PCR) had to learn associated representations just from MeSH definitions and the BERT-CRel objective; their nearest neighbors, hence, tend to be other MeSH codes. However, other acronyms with enough annotation examples in the PubTator dataset (e.g., NO, ROS, AD) mapped to more of the related regular words. Among top five matches for AD and its MeSH code is ``abeta'' (stands for amyloid beta), the main component in plaques in brains of people with Alzheimer's disease.

\subsection{Intrinsic quantitative evaluation}

We now focus on quantitative evaluations based on expert curated datasets  in Table~\ref{tbl:intr-eval-sets}. MiniMayoSRS and Pedersen's datasets are judged by two different groups of experts: physicians and medical coders. We compare our model against several state-of-the-art methods across all the reference datasets.
Table~\ref{tbl:res-final} shows the results of our pre-trained embeddings (fastText$^{+}$) and the fine-tuned embeddings (BERT-CRel). The metric is Spearman's $\rho$ comparing methods' rankings with human relevance scores.
Before we delve into the scores, we note that the correlation coefficients may not be directly comparable in all cases.
Most of the previous studies evaluated the models on a subset of the original reference standards. We specify the number of instances used in each evaluation in  parentheses next to the score; a score without the number of instances means that the evaluation used the full dataset.

\begin{table*}[htb]
\renewcommand{\arraystretch}{1.2}
    \caption{{Results of intrinsic evaluations measured with Spearman's correlation coefficient. Note, the number in parenthesis indicates the number of examples used for the evaluation (with the header row indicating the total number of instances in the original datasets). Scores without parentheses use the full set of instances. $\dagger$ indicates top scores from prior results and our best result used in computing the p-value. The ranking for the word+MeSH rows is computed by the reciprocal rank fusion with the rankings generated by the ``word'' and ``MeSH'' embeddings. }}
    \label{tbl:res-final}
    \centering
    \begin{adjustbox}{width=\textwidth}
    \begin{tabular}{@{}lllllllll@{}}
        \toprule
        Approach &  \specialcell{UMNS\\(n=566)} & \specialcell{UMNR\\(n=587)} & \specialcell{MAYO\\(n=101)} & \specialcell{MMYP\\(n=29)} & \specialcell{MMYC\\(n=29)} & \specialcell{PDSP\\(n=30)} & \specialcell{PDSC\\(n=30)} & \specialcell{HLTK\\(n=36)} \\
        \midrule
        Word2vec (baseline) & 0.568 & 0.499 & 0.508 $\dagger$ & 0.744 & 0.748 & 0.738 & 0.736 & 0.434 \\
        \midrule
        Wang et al.~\cite{wang2018comparison} & 0.440 & n/a & 0.412 & n/a & n/a & 0.632 & n/a & 0.482 \\
        Park et al.~\cite{park2019concept} & n/a & n/a & n/a & n/a & n/a & 0.795 $\dagger$ & n/a & 0.633 $\dagger$ \\
        Chiu et al.~\cite{chiu2016train} & 0.652 (459) & 0.601 (561) & n/a & n/a & n/a & n/a & n/a & n/a \\
        Zhang et al.~\cite{zhang2019biowordvec} & 0.657 (521) & 0.617 (532) & n/a & n/a & n/a & n/a & n/a & n/a \\
        Yu et al.~\cite{yu2016retrofitting,yu2017retrofitting} & 0.689 (526) & 0.624 (543) & n/a & 0.696 (25) & 0.665 (25) & n/a & n/a & n/a \\
        Henry et al.~\cite{henry2019association} & 0.693 (392) $\dagger$ & 0.641 (418) $\dagger$ & n/a & 0.842 $\dagger$ & 0.816 $\dagger$ & n/a & n/a & n/a \\
        \midrule
        fastText$^{+}$ (word) & 0.654 & 0.609 & 0.630 & 0.851 & 0.853 & 0.820 & 0.831 & 0.513 \\
        fastText$^{+}$ (MeSH) & 0.648 & 0.568 & 0.608 & 0.739 & 0.701 & 0.612 & 0.612 & 0.846 $\dagger$ \\
        fastText$^{+}$ (word+MeSH) & 0.689 & 0.623 & 0.685 & 0.836 & 0.832 & 0.756 & 0.769 & 0.753 \\
        BERT-CRel (word)      & 0.683 & 0.643 $\dagger$ & 0.667 & 0.890 $\dagger$ & 0.844 & 0.850 $\dagger$ & 0.849 $\dagger$ & 0.537   \\
        BERT-CRel (MeSH)      & 0.659 & 0.576 & 0.610 & 0.710 & 0.712 & 0.678 & 0.678 & 0.823  \\
        BERT-CRel (word+MeSH) & 0.708 $\dagger$ & 0.637 & 0.695 $\dagger$ & 0.847 & 0.857 $\dagger$ & 0.803 & 0.835 & 0.743  \\ \midrule
        p-value (ours vs SoTA) & 0.328 & 0.479 & 0.0186 & 0.475 & 0.310 & 0.264 & 0.126 & 0.0221\\
        \bottomrule
    \end{tabular}
       \end{adjustbox}
\end{table*}

As indicated in Section~\ref{sec:quant-evals}, we use all instances of all datasets in the evaluation; for any OOV term, we use a fallback mechanism that returns a score either using the \texttt{[UNK]} embedding or the default score $0$. We believe this is a more robust way of evaluating methods instead of selectively ignoring some instances\footnote{In our observation, this was mostly done by other efforts when dealing with terms that are very rare, hence OOV, and hence cannot be readily compared for lack of a proper representation. To some extent, we overcame OOV by using MeSH definitions in fastText$^+$ and the concept pair relevance setup in BERT-CRel}. All rows except those that involve ``MeSH'' in the first column use word-embedding based rankings. Rows that involve MeSH are comparisons that directly compute cosine score with the MeSH code embedding generated by our method. Rows with ``word+MeSH'' modeling involve reciprocal rank fusion~\cite{cormack2009reciprocal} of rankings generated by ``word'' and ``MeSH'' configurations in the previous two rows.

Digging into the scores from Table~\ref{tbl:res-final}, with very few exceptions, BERT-CRel correlates better with human judgments compared with  fastText$^{+}$ across datasets, and improves by around 2.5\% in $\rho$  on average. The most comparable scores with previous efforts are from the third row from the end (BERT-CRel with ``word'' level comparison) given they are word-based measures. This BERT-CRel configuration wins outright for the UMNR dataset even when compared to methods that fuse rankings from word and concept level scores. It also is better than almost all other prior methods across all datasets even when they use selected subsets from the full dataset. 
{
The p-values displayed in the last row for each column were computed use the dagger tagged scores (ours vs prior best).  Except for MAYO and HLTK datasets, our improvements were not statistically significant. 
An important remark here is the rankings (and associated correlation) are not directly comparable in the larger datasets (first two columns of Table}~\ref{tbl:res-final}). 
{As indicated earlier, correlation scores in top scoring prior efforts were generated on smaller datasets where some test term pairs were deliberately left out; nearly a third of the full dataset was ignored in the top scoring study for the larger UMNS and UMNR datasets. So even though the Fisher Z-transformation method we used accounts for varying sample sizes, due to the way the smaller samples were curated (by selectively, \textbf{not} randomly, eliminating certain rare term pairs), our results are more robust for  larger datasets despite these findings regarding statistical significance.}

\subsection{Extrinsic quantitative evaluation}
We further investigate the efficacy of the fine-tuning process using a semi-supervised learning method in a simple downstream NLP task setup. We use the same annotation dataset (i.e., PubTator) used for the joint learning method for the biomedical entity linking (EL) problem in this evaluation. The EL goal is to disambiguate the associations between mentions (entity spans) and the unique entity identifiers (MeSH codes). In this evaluation, a MeSH code is represented by the mean vector of the constituent word embeddings for the NLM defined entity name or more directly the MeSH code embedding. We rank all MeSH codes given a mention phrase using cosine similarity score of the average embedding vector of the mention phrase and the corresponding average for MeSH code's preferred name (rows 1 and 3 of Table~\ref{tbl:rst-en}); if the similarity is computed using the vectors for MeSH codes, we obtain the MeSH ranking. The word+MeSH ranking (rows 2 and 4 of Table~\ref{tbl:rst-en}) is the ranking based on RRF fusion of word and MeSH based rankings. As the table shows the BERT-CRel fine-tuning consistently improves EL results.

\begin{table}[hbt]
  \centering
  \caption{{Results of the entity linking task with the pre-trained and fine-tuned static embeddings (Number of test examples: 21,505 mention-entity pairs)}}
  \label{tbl:rst-en}
  \begin{tabular}{@{}lcc@{}}
    \toprule
    & Top-1 accuracy & Top-5 accuracy \\
    \midrule
    fastText$^{+}$ (word) & 0.357 & 0.512 \\
    fastText$^{+}$ (word+MeSH) & 0.442 & 0.703 \\
    BERT-CRel (word) & 0.368 & 0.530 \\
    BERT-CRel (word+MeSH) & 0.474 & 0.741 \\
    \bottomrule
  \end{tabular}
\end{table}

Our effort provides the most robust evaluation  by exhaustively considering all instances across all well-known datasets developed for evaluating embeddings. Overall, we demonstrate that jointly learning word and concept embeddings by leveraging definitional information for concepts provides better embeddings; further enhancing these embeddings by exploiting distributional correlations across concepts (obtained from MeSH co-occurrences and hierarchical links),  through transformer-based classifiers, offers more noticeable gains in embedding quality.

\subsection{Limitations and future directions}
{Our work can be improved in a few directions that also indicate some of the limitations of this effort. Although our methods are novel and they helped us improve embedding quality, the performance gains are clearly not spectacular; even when the gains are substantial for smaller datasets, they were not statistically significant. Training on much bigger citation subsets both for fastText and the BERT-CRel fine-tuning may result in further improvements. Also, expanding beyond MeSH and considering other relevant vocabularies (e.g., SNOMED-CT, ICD-10) may help. For example, the University of Kentucky medical center has over 15 million patient visits to its clinics over the past decade. Each patient visit is assigned a set of ICD-10 codes  by a trained coder (just like MeSH terms assigned to a PubMed article). In our prior work,  we used correlations in these ICD-10 code sets to improve automatic electronic medical record (EMR) coding efforts}~\cite{kavuluru2015empirical}. {BERT-CRel can use disease concept pairs derived from EMRs to further fine-tune word embeddings. Next, in terms of sampling PubMed citations either for fastText pre-training or for fine-tuning with concept pairs, we chose a random order. Though this works in general, for rare words and concepts, this simple strategy may not lead to high quality embeddings for them. Although we addressed this to some extent with MeSH definitional information, a more targeted heuristic that over-samples citations that have been tagged with rare concepts may naturally lead to better representations for them and associated words that describe them.}

\section{Conclusion~\label{sec:conclusions}}
In this effort, we proposed a method for training and improving static embeddings for both words and domain-specific concepts using a neural model for the concept-relatedness classification task. To incorporate the relational information among biomedical concepts, we utilize document metadata (i.e., MeSH assignments to the PubMed articles) in corpus and the hierarchical relationships of the concepts defined in a controlled vocabulary (i.e., MeSH hierarchy structures). Our approach achieved the best performances across several benchmarks. Qualitative observations indicate that our methods may be able to nudge embeddings to capture more precise connections among biomedical terms.

Our proposed method for training and improving static embeddings can be utilized in many BioNLP tasks. The use of joint word/concept embeddings can potentially benefit neural models that need mutual retrievability between multiple embeddings spaces. In one of our recent studies, we leveraged embeddings generated with these methods in a neural text summarization model for information retrieval~\cite{noh2020literature}. Exploiting the joint embeddings of words and MeSH codes, we were able to summarize a document into a sequence of keywords using either regular English words or MeSH codes that are then compared with query words and codes. We will continue to explore applications of these embeddings in other future applications in knowledge discovery and information retrieval. Other researchers can use them in their own tasks by downloading them from our publicly available repository: \url{https://github.com/bionlproc/BERT-CRel-Embeddings}

\section*{Acknowledgements}
Research reported in this publication was supported by the National Library of Medicine of the U.S.~National Institutes of Health under Award Number R01LM013240. The content is solely the responsibility of the authors and does not necessarily represent the official views of the National Institutes of Health.

\bibliographystyle{ieeetr}
\bibliography{JBI-ImprovStaticEmbed-revision.bib}

\end{document}